\documentclass[runningheads]{llncs}
\usepackage{lipsum}
\usepackage{xcolor}
\usepackage{booktabs}
\usepackage{graphicx}
\usepackage{hyperref}
\usepackage{manyfoot}
\usepackage{subfigure}
\usepackage[T1]{fontenc}
\usepackage{amsmath,amssymb,amsfonts}

\usepackage[ruled,linesnumbered]{algorithm2e}
\usepackage{comment}
\usepackage[flushleft]{threeparttable}
\begin{document}
\title{When Molecular GAN Meets Byte-Pair Encoding}

\author{Huidong Tang \inst{1,3} 
\and Chen Li* \inst{2} 
\and Yasuhiko Morimoto \inst{3}
}

\authorrunning{H. Tang et al.}

\institute{Shandong Xiehe University, Jinan, China\\
\email{tanghd24@163.com} \\ \and
Graduate School of Informatics, Nagoya University, Nagoya, Japan \\
\email{li.chen.z2@a.mail.nagoya-u.ac.jp} \\ \and 
Graduate School of Advanced Science and Engineering, Hiroshima University, Higashi-Hiroshima, Japan \\
\email{morimo@hiroshima-u.ac.jp}
}

\maketitle  

\begin{abstract}
Deep generative models, such as generative adversarial networks (GANs), are pivotal in discovering novel drug-like candidates via \textit{de novo} molecular generation. However, traditional character-wise tokenizers often struggle with identifying novel and complex sub-structures in molecular data. In contrast, alternative tokenization methods have demonstrated superior performance. This study introduces a molecular GAN that integrates a byte level byte-pair encoding tokenizer and employs reinforcement learning to enhance \textit{de novo} molecular generation. Specifically, the generator functions as an actor, producing SMILES strings, while the discriminator acts as a critic, evaluating their quality. Our molecular GAN also integrates innovative reward mechanisms aimed at improving computational efficiency. Experimental results assessing validity, uniqueness, novelty, and diversity, complemented by detailed visualization analysis, robustly demonstrate the effectiveness of our GAN.
\keywords{Molecular generation \and Byte-pair encoding \and GANs.}
\end{abstract}

\section{Introduction}
\label{sec:introduction}
\begin{figure}[t]
\centering
\includegraphics[width=1\textwidth]{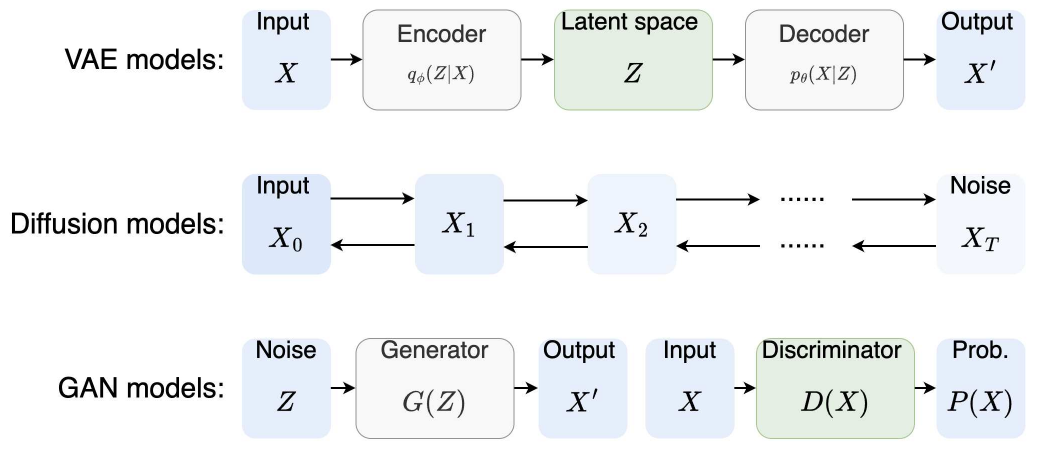}
\caption{Overview of three deep generative models.}
\label{fig:deep_generative_models}
\end{figure}

\textit{De novo} molecular generation is a pivotal pursuit in drug discovery, aimed at identifying novel candidate molecules that exhibit desirable drug-like properties and are easy to synthesize \cite{meyers2021novo,martinelli2022generative}. However, exploring vast chemical spaces to uncover these molecules is acknowledged for its inherent challenges, being both time-consuming and costly \cite{medina2014balancing}. In response to these challenges, artificial intelligence (AI) methodologies \cite{zhang2024cd,zhang2024quantitative,zhang2020hierarchy,zhang2019predicting}, especially deep generative models such as variational autoencoders (VAEs) \cite{kusner2017grammar,ma2021gf}, diffusion-denosing models \cite{vignacdigress,xu2023geometric}, and generative adversarial networks (GANs) \cite{li2022transformer,li2023spotgan}, have emerged as powerful tools for designing and refining chemical candidates. These models, illustrated in Figure \ref{fig:deep_generative_models}, employ deep learning techniques to generate molecular structures, accelerating the discovery of drug candidates in a manner that is both efficient and cost-effective.

Typically, molecules in a GAN can be represented using either 1-dimensional simplified molecular-input line-entry system (SMILES) strings \cite{weininger1988smiles} or molecular graphs \cite{kearnes2016molecular}. SMILES strings are preferred for their straightforward representation and sequential structure, akin to sequences in natural language processing (NLP) \cite{chowdhary2020natural}. While the simplicity of SMILES strings is beneficial for eliminating redundant information and providing a compact input representation, it represents a high-level abstraction of molecular structures. Such abstraction compresses high-dimensional information into a low-dimensional format. The intricate relationships between atoms may be lost when using traditional character-level tokenizers, which divide molecules into atomic symbols, integers for ring closure, and symbols for bonds and chirality. To better preserve these hidden relationships, we propose byte-level byte-pair encoding (BPE) tokenization \cite{sennrich2015neural}, as utilized in NLP \cite{liu2019roberta,radford2019language}. BPE tokenization generates a vocabulary based on frequent sub-structures, thereby maintaining these subtle relationships.

The representation of SMILES strings poses several challenges in GAN training. Firstly, the discrete nature of SMILES strings introduces unique obstacles, such as gradient vanishing during training, which can impede the convergence and effectiveness of traditional GAN approaches \cite{guimaraes2017objective}. Secondly, each character denotes specific atoms and bonds, necessitating careful handling of discrete outputs throughout the training process \cite{li2024tengan}. This discreteness affects the smoothness of the optimization landscape, potentially leading to difficulties in learning molecular representations.

To tackle these challenges, researchers often employ specialized techniques such as reinforcement learning (RL) \cite{williams1992simple} or innovative loss functions tailored to discrete outputs. These methods aim to stabilize training and enhance the ability of GANs to generate realistic and diverse molecular structures. In this study, we introduce an innovative molecular GAN that utilizes BPE tokenizer and integrates RL algorithms for \textit{de novo} molecular generation. Initially, our GAN employs the BPE tokenizer for tokenization, and generating embeddings of SMILES strings which serve as inputs for both the generator and discriminator. The generator, implemented as an long short-term memory (LSTM) \cite{graves2012long}, and the discriminator, as a bidirectional LSTM, autonomously learn semantic and syntactic aspects from these embeddings. To stabilize training, we employ a RL algorithm that effectively balances exploration and exploitation during learning phases. The generator acts as an actor, generating SMILES strings, while the discriminator serves as a critic, evaluating their quality. Operating with autoregressive predictions at the atomic level, the generator iteratively builds upon previously generated atoms, while the discriminator assesses atoms bidirectionally, including the current prediction. Additionally, our molecular GAN incorporates innovative reward mechanisms designed to optimize computational efficiency. These mechanisms aim to improve overall performance and reliability in molecular generation tasks. The main contributions are highlighted as follows:

\begin{itemize}
\item {\bf Integration of BPE tokenizer}: We integrate a BPE tokenizer to tokenize SMILES strings and generate embeddings, preserving the hidden relationships between atoms.

\item {\bf GAN stability and innovative reward mechanisms}: Implementation of a RL algorithm stabilizes the training of our GAN, addressing challenges like gradient vanishing. Our GAN also integrates novel reward mechanisms to enhance computational efficiency. These advancements aim to elevate performance in molecular generation.

\item {\bf Comprehensive experimental evaluation}: Extensive experimental assessments are conducted covering validity, uniqueness, and novelty metrics, supported by visualization analysis. These evaluations were crucial in assessing the effectiveness and robustness of our GAN in molecular generation.
\end{itemize}

\section{Related Work}
\label{sec:related}

\subsection{Sequential Molecular Representations}
\label{sec:1d_rep}
Recent research has witnessed VAEs such as characterVAE \cite{kusner2017grammar}, GrammarVAE \cite{kusner2017grammar}, and GxVAEs \cite{li2024gxvaes} employing an encoder-decoder architecture for molecular generation. The encoder convert SMILES strings into a latent space, while the decoder generates new molecules. VAEs are effective in exploring chemical space and generating diverse molecular structures. SyntaxVAE \cite{dai2018syntax} incorporates attribute grammar to enforce semantic validity during parse-tree generation.

On the other hand, GAN models such as MacGAN \cite{tang2023macgan} operate with a generator network that produces new molecules and a discriminator network trained to differentiate real molecules from their SMILES representations. Due to the discrete nature of SMILES strings, many GAN models employ RL algorithms to enhance stability and effectiveness. For example, TransORGAN \cite{li2022transformer} and TenGAN \cite{li2024tengan} employ the Monte Carlo tree search RL algorithm \cite{browne2012survey} to compute rewards from the discriminator. These rewards guide their generators in generating molecules either from scratch \cite{li2024tengan} or based on given scaffolds \cite{li2023spotgan}. However, the computational cost associated with the sampling phase limits their scalability to large chemical databases. In contrast, EarlGAN \cite{tang2023earlgan} introduces a RL algorithm \cite{bahdanau2022actor} for molecular generation from SMILES strings. It balances the trade-off between generation accuracy and sampling efficiency.

Self-referencing embedded strings (SELFIES) \cite{krenn2020self}, another 1-dimensional sequential representation, have emerged as a notable method for molecular representation. It encodes molecular structures using strings of symbols. It offers efficient representation and manipulation of chemical structures, supported by a comprehensive library for translation between SMILES strings and SELFIES representations. SELFIES representations ensure the generation of valid molecules and are particularly beneficial for tasks like de novo molecule design and property optimization. However, SELFIES representations are limited in fully representing complex macromolecules and crystals that consist of large molecules or intricate bonding patterns \cite{krenn2022selfies}. In this study, our primary focus revolves around molecular generation from SMILES strings. This approach is chosen due to the widespread adoption and well-established methodologies within deep generative models for molecular generation.

\subsection{Graphical Molecular Representations}
\label{sec:2d_rep}
Graph-based approaches represent molecules as graphs, where atoms serve as nodes and bonds as edges. MolGAN \cite{de2018molgan} utilizes a discrete GAN framework to generate molecular graphs, focusing on sampled atomic and chemical bond features to maximize likelihood. The model integrates a reward mechanism to assess the quality of generated molecular structures. Despite these advancements, MolGAN faces challenges with mode collapse, a common issue in generative models that limits diversity in generated molecular structures. ALMGIG \cite{polsterl2021adversarial}, an extension of the bidirectional GAN, enhances molecular generation by learning molecular space distributions through adversarial cyclic consistency loss. GraphAF \cite{shigraphaf} is a flow-based autoregressive model designed for molecular graph generation, integrating autoregressive and flow-based approaches \cite{dinh2016density} to model real-world molecular data density effectively. It utilizes an autoregressive sampling method to sequentially generate nodes and edges based on existing sub-graph structures, ensuring the validity of generated molecular graphs by adhering to chemical domain knowledge and valency rules. Unlike other models like  graph convolutional policy network \cite{you2018graph} and molecular recurrent neural network \cite{popova2019molecularrnn}, GraphAF employs a feedforward neural network during training to compute the exact data likelihood efficiently in parallel. 

\subsection{Tokenization Approaches for Molecular Generation}
\label{sec:tokenization}
Tokenization is a critical preprocessing step in NLP that involves breaking down text into smaller, manageable units called tokens. This process is essential for enabling computers to process and analyze text data effectively. In recent years, tokenization has seen significant advancements, particularly with the advent of pre-trained language models such as BERT \cite{devlin-etal-2019-bert} and GPT \cite{achiam2023gpt}, which have greatly enhanced the capabilities of various applications. For example, character-level tokenization has been effectively employed in the field of molecular design. One notable approach is SpotGAN \cite{li2023spotgan}, which tokenizes molecular scaffolds and uses the MCTS algorithm to predict molecular decorations. Similarly, TransORGAN \cite{li2022transformer} leverages transformer architecture to capture and interpret semantic meanings from tokenized SMILES strings. Additional innovations in tokenization have led to the development of specialized methods tailored to the molecular domain. SMILES pair encoding (SPE) \cite{li2021smiles}, for example, utilizes BPE in the context of SMILES strings, integrating it with existing machine learning models to enhance performance. Atom-in-SMILES (AIS) \cite{ucak2023improving} introduces molecular environment information around atoms into the tokenization process, which helps uncover hidden relationships within molecules.

Despite these advancements, many of these specialized tokenization methods still fall short in addressing the intricate connections between tokenized SMILES and their respective models. Our model aims to bridge this gap by incorporating an efficient reward mechanism, which reduces computational complexity while accommodating an expanded vocabulary through BPE tokenization. This approach not only streamlines the tokenization process but also enhances the overall efficiency and effectiveness of molecular generation models.

\section{Model Description}
\label{sec:model}
\begin{figure}[ht]
\centering
\includegraphics[width=1\textwidth]{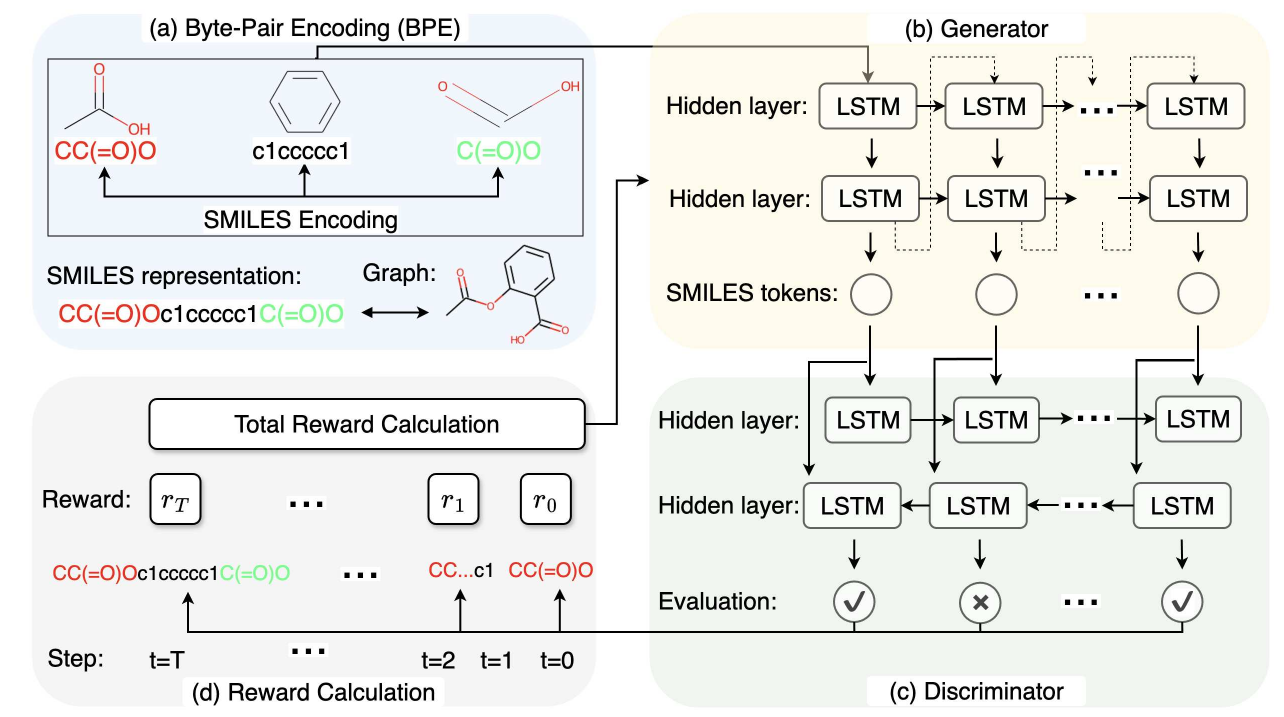}
\caption{Overview of the architecture of our molecular GAN.}
\label{fig:overview}
\end{figure}
Figure \ref{fig:overview} provides an overview of our proposed molecular GAN. Molecules in SMILES representations are initially tokenized using a BPE tokenizer (depicted in Figure \ref{fig:overview} (a)). These SMILES tokenizations serve as inputs for both the generator (illustrated as an LSTM in Figure \ref{fig:overview} (b)) and the discriminator (represented by a bidirectional LSTM in Figure \ref{fig:overview} (c)). The generator is then trained using a RL algorithm, guided by the discriminator (Figure \ref{fig:overview} (d)).

\subsection{SMILES Tokenization} 
BPE is a compression algorithm that combines both character-level and word-level representations. It works by iteratively and greedily identifying and merging the most frequent pairs of characters or subwords, which helps decompose rare or unknown words into more common, smaller subword units \cite{Kida1999BytePE}. This process continues until the desired vocabulary size is achieved, allowing the representation of an entire corpus with a manageable set of subword units. The detailed BPE algorithm can be found under Algorithm \ref{alg:bpe}.

\SetKwInOut{KwIn}{Initialization}
\begin{algorithm}[t] 
\caption{Byte-Pair Encoding Algorithm.}
\label{alg:bpe}
\KwData{a SMILES dataset $S_{r}=\{X\}$, a vocabulary size $V$}
\KwIn{the vocabulary with all byte tokens in the SMILES corpus with initial vocabulary size $V_{intial}$}

\tcp{Frequency Calculation}
Compute the frequency of each byte token. \\
\For{$v=V_{intial}$ {\bfseries to} vocabulary size $V$}{
Find the most frequent token pair. \\ 
Merge the most frequent token pair for a new token. \\ 
Update the frequent token pair list. \\ 
Add the new token into the vocabulary. \\ 
}
\end{algorithm}

\subsection{Value Function of Molecular GAN}
\label{sec:gan}
When applying RL algorithms in GANs, a common challenge emerges when rewards are calculated across the entire sequence early in training, leading to an imbalance in the competition between the generator and discriminator. To address this issue, our molecular GAN employs a strategy where a reward is assigned to each token, ensuring more consistent feedback from the discriminator. The corresponding value function of our GAN is formulated as follows:
\begin{align}
\min_\theta\max_\phi V (G_\theta, D_\phi) \nonumber =  &\sum_{t=1}^{T}\mathbb{E}_{p_{data}(x_t|\overrightarrow{x}_{1:t-1})}[\log D_\phi (y_t|\overrightarrow{x}_{1:t}, \overleftarrow{x}_{t:T})] \nonumber\\
& + \sum_{t=1}^{T}\mathbb{E}_{p_\theta(x_t|\overrightarrow{x}_{1:t-1})}[\log (1-D_\phi (y_t|\overrightarrow{x}_{1:t}, \overleftarrow{x}_{t:T})].
\end{align}
Here, $y_t$ denotes the discriminator's prediction of atom $x_t$, based on both the forward pass $\overrightarrow{x}_{1:t}$ and the backward pass $\overleftarrow{x}_{t:T}$ over the atoms. $T$ signifies the maximum length of SMILES strings. 

\subsection{Reward Calculation for Molecular GAN}
Moreover, we adopt innovative reward strategies instead of cumulative rewards for efficiency, tailored to the characteristics of SMILES strings. The reward $R_t$ is defined as follows:
\begin{align}
&R_t = 2r_t - 1, \\
&r_t = D_\phi(\hat{y}_t|\overrightarrow{\hat{x}}_{1:t},\overleftarrow{\hat{x}}_{t:T}), 
\end{align}
where $\hat{y}_t$ signifies the prediction for the generated atom $\hat{x}_t$ by the discriminator. $\overrightarrow{\hat{x}}_{1:t}$ and $\overleftarrow{\hat{x}}_{t:T}$ represent the forward pass and backward pass over the generated atoms, respectively. These rewards are represented as $2r_t-1$. 

\subsection{Loss Function}
Additionally, our molecular GAN integrates a baseline to mitigate variance and highlight the relative reward of each atom, akin to standard RL practices. The baseline in our GAN is derived from an exponential moving average of the batch's mean reward. The generator's loss function is as follows:
\begin{align}
\text{Loss} = & - \frac{\sum_{n=1}^{N}\sum_{t=1}^{T_n}(R^n_t - b_i)\log p_\theta(\hat{x}^n_t|\overrightarrow{\hat{x}}^n_{1:t-1})}{\sum_{n=1}^{N}T_n},
\end{align}
where $N$ denotes the batch size, $T_n$ represents the length of the $n$-th molecule's string. Here, $b_i$ denotes the current baseline, computed as follows:
\begin{align}
&b_i = \alpha b_{i-1} + (1-\alpha)\hat{R}_{i},
\end{align}
where $\hat{R}_{i}$ represents the average reward of the atoms within the current batch. The parameter $\alpha$ is fixed at $0.9$, and $b_i$ is updated on a per-batch basis. 

\SetKwInOut{KwIn}{Initialization}
\begin{algorithm}[t] 
\caption{Training Procedure for our molecular GAN.}
\label{alg:gan}
\KwData{a SMILES dataset $S_{r}=\{X\}$}
\KwIn{the generator $G_{\theta}$, discriminator $D_{\phi}$}

\tcp{SMILES tokenization}
Tokenize the SMILES strings in the training dataset. \\

\For{$i=0$ {\bfseries to} maximum steps}{
Input noise to the generator $G_{\theta}$. \\ 
The generator $G_{\theta}$ produces SMILES strings $S_{generated}$ by sequentially generating each token autoregressively.\\

\tcp{Train the discriminator}
Input the real dataset $S_r$ and the generated data $S_f$ to the discriminator $D_{\phi}$. \\
The discriminator $D_{\phi}$ is trained on real $S_r$ and the generated data $S_f$.\\
Update $\phi$ of the discriminator $D_{\phi}$. \\

\tcp{Train the generator}
The trained discriminator $D_{\phi}$ discriminates $S_f$. \\
Calculate the reward $R_t$ for each token based on the discrimination. \\
Update $\theta$ of the generator $G_{\theta}$. \\
}	
\end{algorithm}

Algorithm \ref{alg:gan} outlines the primary training procedure of our molecular GAN. Initially, SMILES strings representing molecules in the training dataset are tokenized by BPE tokenizer. Subsequently, our molecular GAN leverages a RL algorithm. The generator acts as the actor, responsible for producing SMILES strings, while the discriminator acts as the critic, evaluating the quality of generated samples. This dual-role framework enables our GAN to iteratively improve the quality of generated molecules. The discriminator's assessments, manifested as rewards, play a crucial role in guiding the generator's updates. By leveraging these rewards, our GAN fine-tunes its parameters to optimize the generation of molecules with novel structures.
\section{Experiments}
\label{sec:exp}

\subsection{Dataset}
\label{sec:dataset}
We assessed our molecular GAN using the ZINC dataset \cite{irwin2012zinc}, a prominent benchmark in computational chemistry. With 250,000 drug-like molecules, this dataset offers a comprehensive basis for evaluating our model's effectiveness.

\subsection{Evaluation Metrics}
\label{sec:metrics}

\noindent\textbf{Statistic Metrics.} We assessed our molecular GAN using the statistical metrics of validity, uniqueness, and novelty, and diversity, as outlined in \cite{li2023spotgan,tang2023earlgan}. 
\begin{itemize}

\item \textbf{Validity} indicates the ratio of generated SMILES strings that are syntactically valid according to the rules of molecular structure.

\item \textbf{Uniqueness} measures the proportion of unique molecular structures among all valid molecules generated. This metric is crucial because generating diverse and non-redundant molecules helps in exploring a broader chemical space and identifying candidates that possess distinct structures.

\item \textbf{Novelty} quantifies the ratio of molecules generated that are not present in the training dataset used to train the generative model.
\item \textbf{Diversity} assesses the structural variety or distinctiveness of the generated molecules. It is often evaluated using metrics such as Tanimoto similarity, which measures the similarity between molecular fingerprints \cite{li2024gxvaes}. Higher diversity scores indicate a wider exploration of chemical space, potentially leading to the discovery of molecules with diverse biological activities.
\end{itemize}  

\noindent\textbf{Chemical Properties.} We evaluated property distributions (drug-likeness, solubility and synthesizability scores) of molecules generated by our GAN.
\begin{itemize}
\item \textbf{Drug-likeness (QED)} scores assess how closely a molecule's properties align with those typically found in drugs.
\item \textbf{Solubility (logP)} scores assess hydrophilicity of a molecule, quantified by the logarithm of the octanol-water partition coefficient. 
\item \textbf{Synthesizability (SA)} scores evaluate the ease or feasibility of synthesizing a molecule in a laboratory setting.
\end{itemize}  

All statistical metrics and chemical properties are scaled to a range of 0 to 1, ensuring consistency and enabling standardized comparisons and interpretations.

\subsection{Hyperparameter Setting}
\label{sec:hyper}
The generator utilizes LSTM as the foundational architecture. The generator consists of input transformation layers for the noise input, one LSTM cell for the extracted feature input, and output layers, with input and output layers featuring a dropout probability of 0.1 \cite{srivastava2014dropout}. Gradient clipping is applied within the range [-0.1, 0.1]. The discriminator employs a bidirectional LSTM layer to process tokenized embeddings, output layers with a dropout rate of 0.1, gradient clipping between -0.1 and 0.1, and L2 regularization with a coefficient of 1e-6. Both the generator and discriminator utilize the Adam optimizer with a learning rate of 5e-6 and the batch size is set to 256.

\subsection{Evaluation Results}
\label{sec:res}

\begin{table}[t]
\setlength\tabcolsep{4pt}
\centering
\caption{Comparisons of our molecular GAN with baselines on the ZINC dataset.}
\label{tab:results}
\begin{tabular}{l|ccccc}\toprule
Model &  Validity (\%) $\uparrow$ & Uniqueness (\%) $\uparrow$ & Novelty (\%) $\uparrow$ & Diversity $\uparrow$  \\\hline
CharacterVAE \cite{kusner2017grammar} & 73.34 & 99.18 & \bf 100.00 & 0.39 \\
GrammarVAE \cite{kusner2017grammar} & 76.36 & 99.55 & \bf 100.00 & 0.45 \\
JTVAE \cite{jin2018junction} & \bf 100.00 & 13.94 & 99.43 & 0.61 \\
GraphAF$^{1\star}$ \cite{shigraphaf} & 72.26 & 84.80 & \bf 100.00 & 0.79 \\
GraphAF$^{10\star}$ \cite{shigraphaf} & 67.27 & 99.44 & \bf 100.00 & 0.69 \\
TransORGAN \cite{li2022transformer} & 75.52 & 94.64 & \bf 100.00 & 0.68 \\
\textbf{Ours} & 89.70 & \bf 99.84 & 99.80 & \bf 0.91 \\\bottomrule
\end{tabular}
\begin{tablenotes}
    \footnotesize
    \item{$\star$} 1 and 10 represent the minimum SMILES string length.
\end{tablenotes}
\end{table}

To assess the efficacy of our molecular GAN, we conducted comparisons with several baseline models: VAE-based CharacterVAE, GrammarVAE \cite{kusner2017grammar}, and JTVAE \cite{jin2018junction}; and graph-based GraphAF \cite{shigraphaf}; and GAN-based TransORGAN \cite{li2022transformer}. These baseline results are cited from the literature \cite{li2022transformer}.

The evaluation results are presented in Table \ref{tab:results}. Note that values in bold represent the maximum values. For the validity metric, JTVAE scored the highest value (100.00\%), and our model comes in second (89.70\%). Our model exhibits the highest uniqueness (99.84\%) and diversity (0.91), due to the BPE tokenization, which preserves the hidden relationships between atoms. These hidden relationships provide diverse sub-structures for the generated molecules, resulting in these results. For the novelty, our model can not reach 100\%, but can be comparable. In conclusion, our model achieves balanced performance at these four metrics, proving its effectiveness in molecular generation.

\begin{figure}[t]
\centering
\subfigure[Drug-likeness scores.]{
\includegraphics[width=0.5\textwidth]{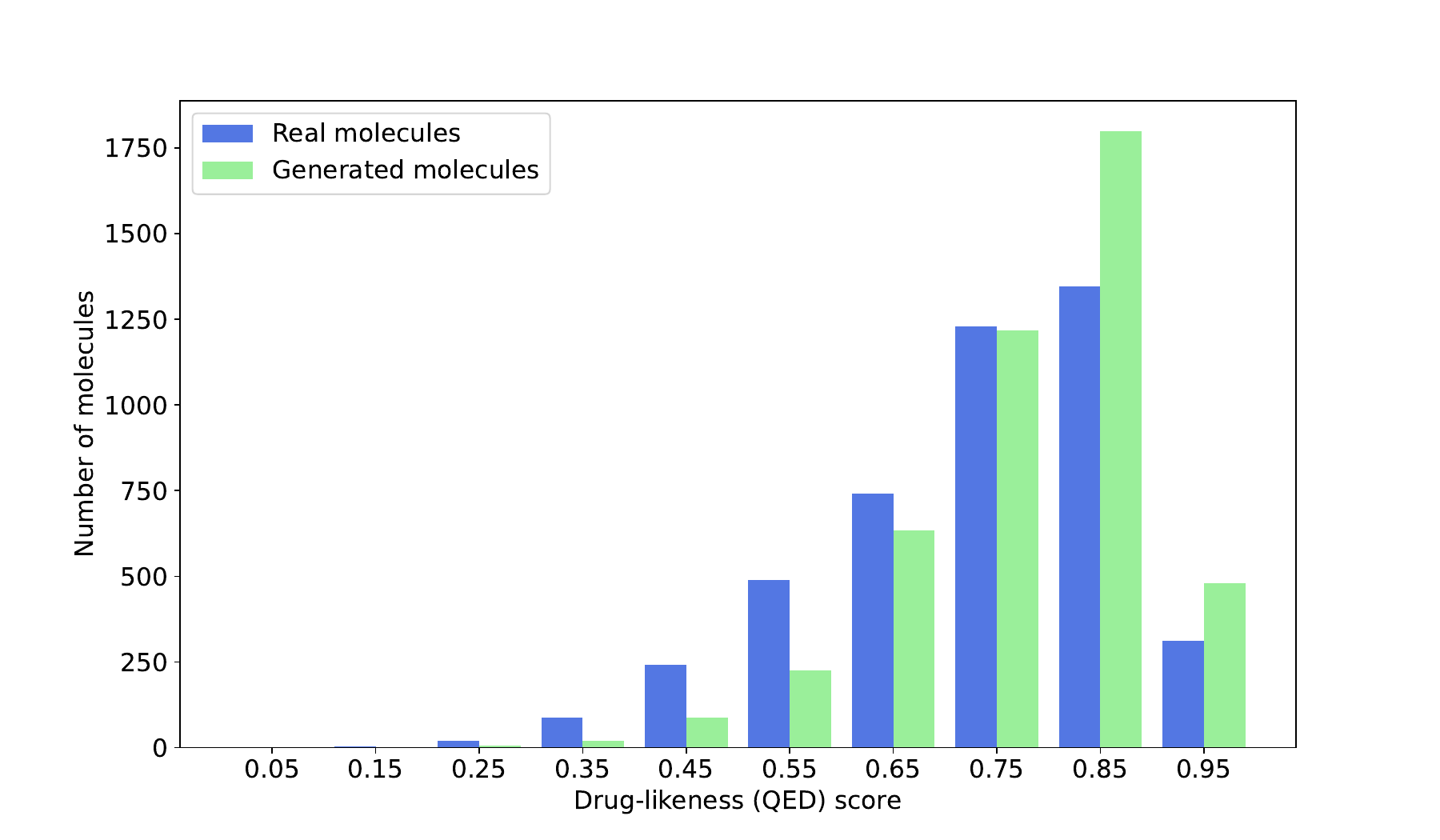}
}\hspace{-8mm}
\subfigure[Solubility scores.]{
\includegraphics[width=0.5\textwidth]{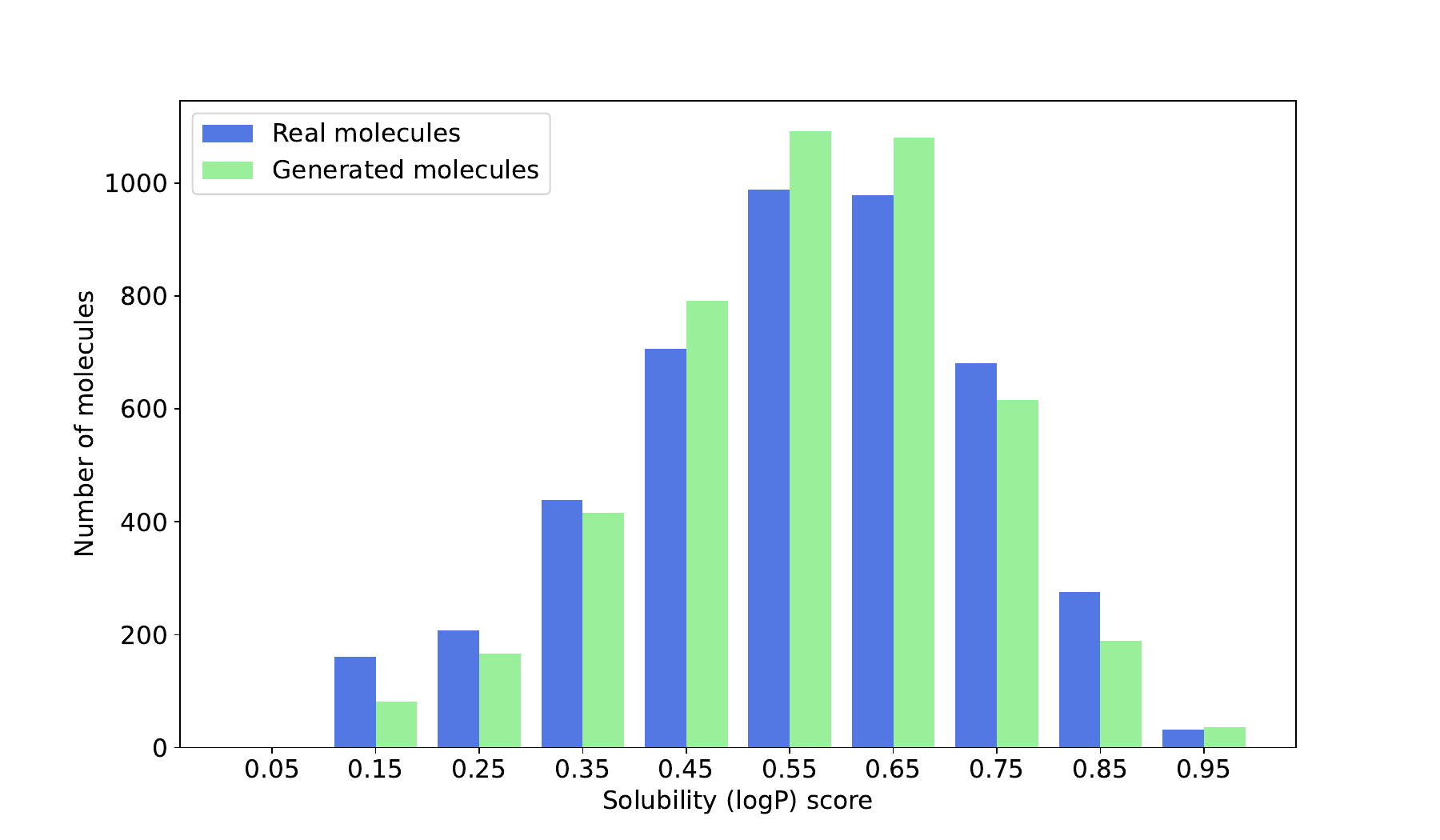}
}\hspace{-8mm}
\subfigure[Synthesizability scores.]{
\includegraphics[width=0.5\textwidth]{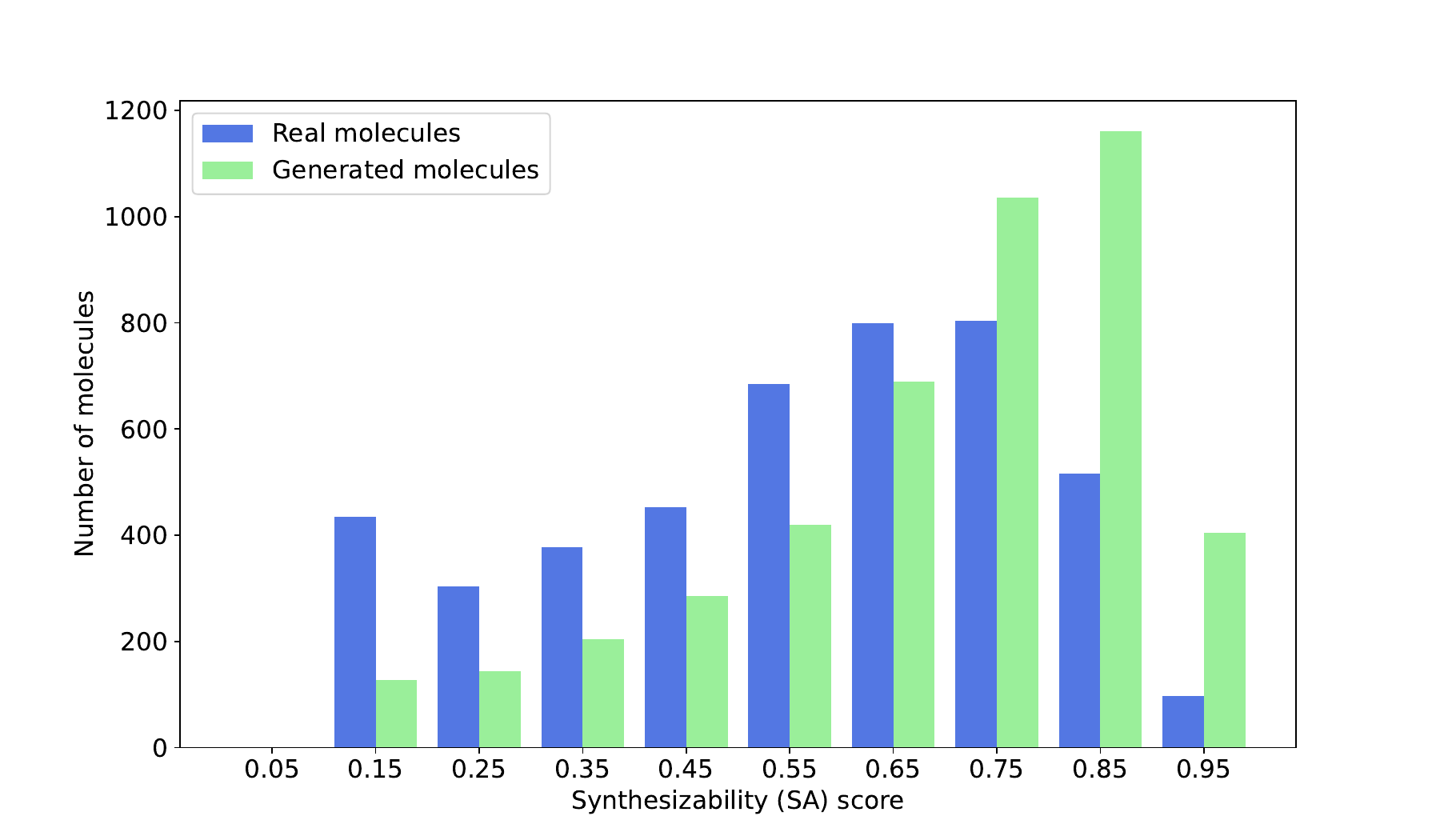}
}
\caption{Distributions of chemical properties of molecules generated by our GAN.}
\label{fig:dist}
\end{figure}

\begin{figure}[ht]
\centering
\subfigure[Drug-likeness scores.]{
\includegraphics[width=0.48\textwidth]{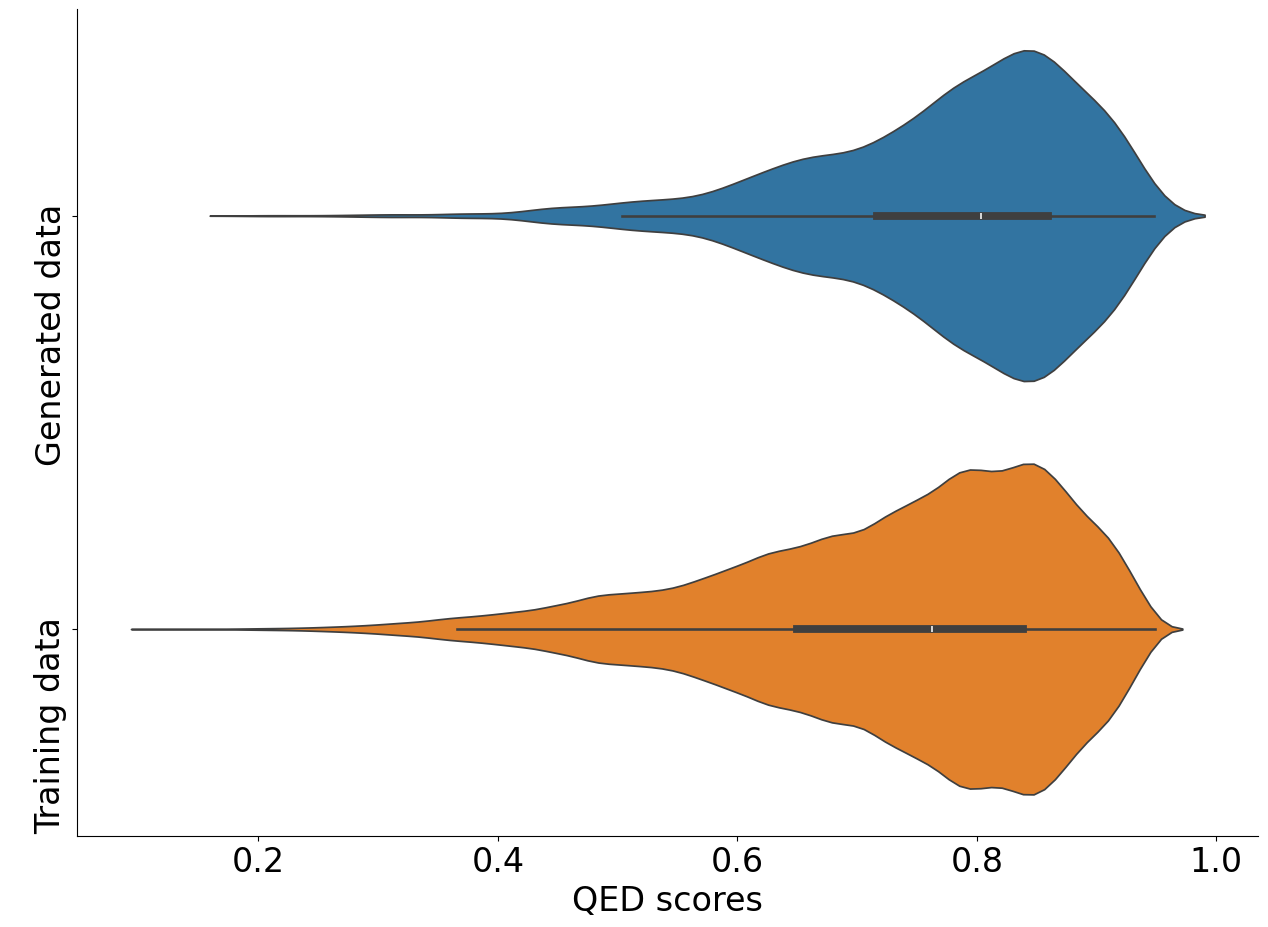}
}\hspace{-4mm}
\subfigure[Solubility scores.]{
\includegraphics[width=0.48\textwidth]{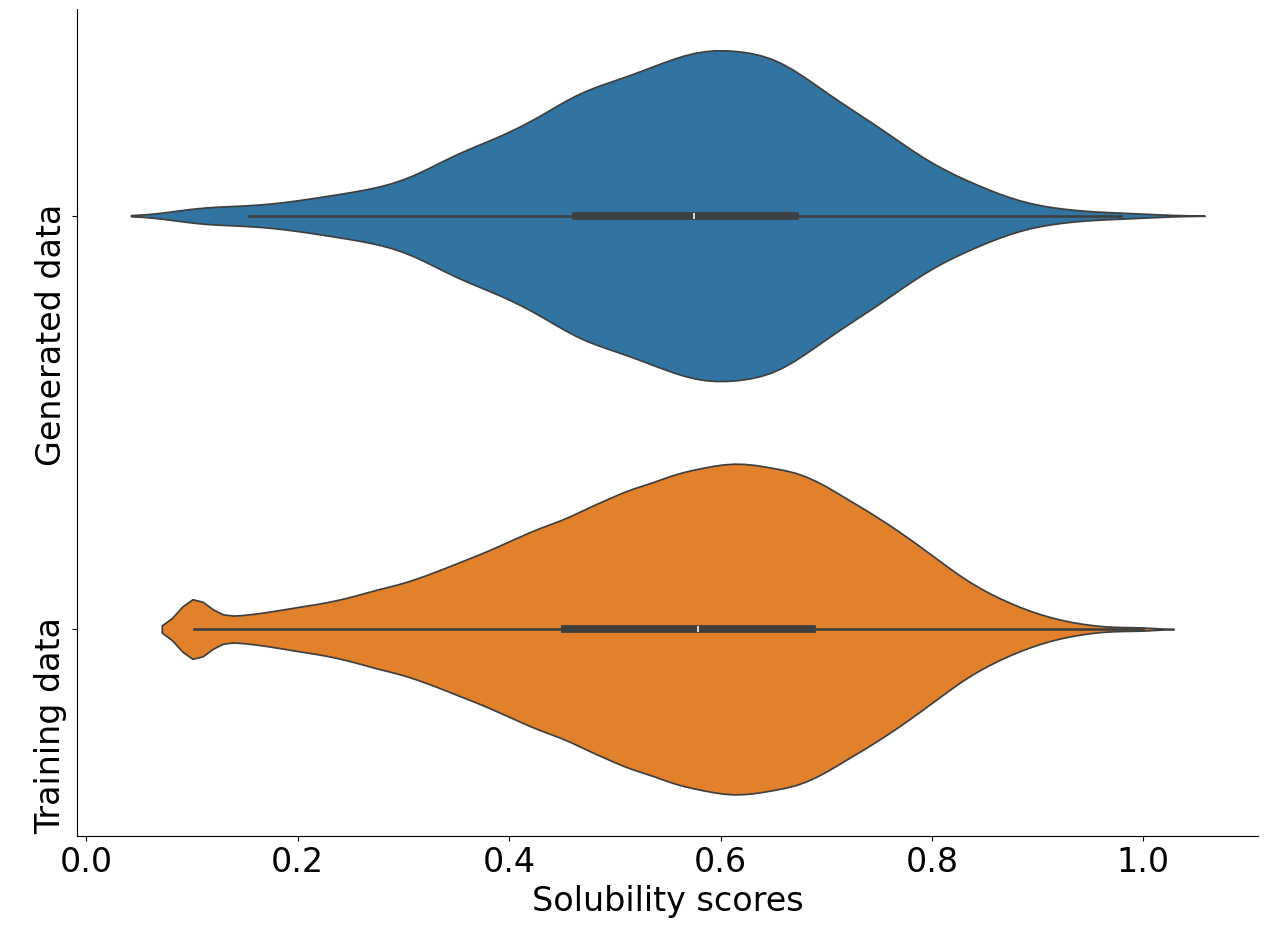}
}\hspace{-4mm}
\subfigure[Synthesizability scores.]{
\includegraphics[width=0.48\textwidth]{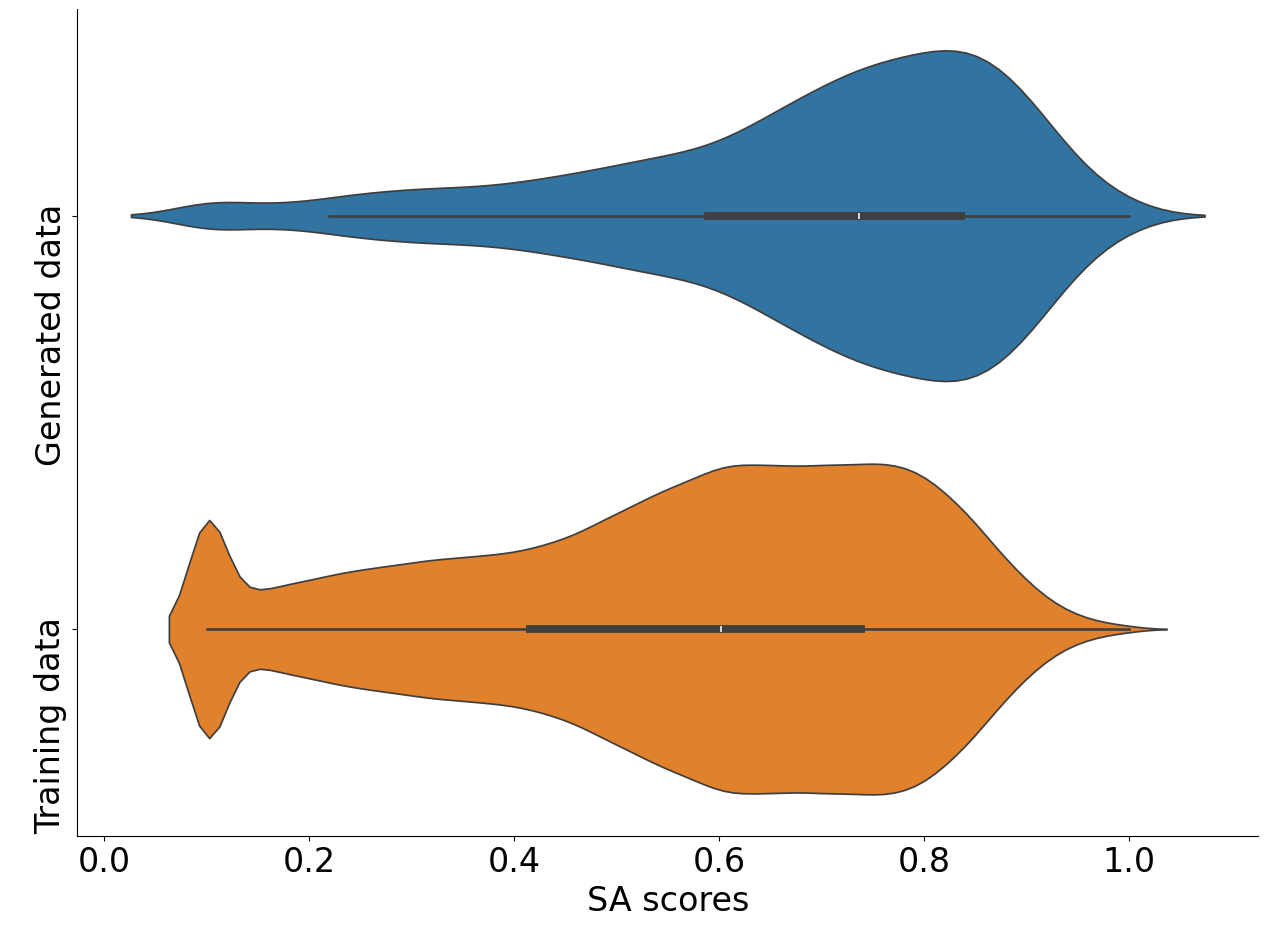}
}
\caption{Violin Plots of chemical properties of molecules generated by our GAN.}
\label{fig:violin}
\end{figure}

\begin{figure}[t]
\centering
\includegraphics[width=0.7\textwidth]{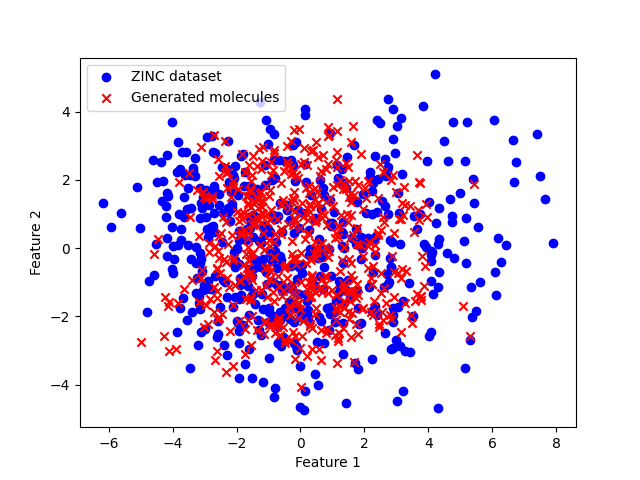}
\caption{Visualization of clustering generated and trained molecules in two-dimensional (2D) space.}
\label{fig:2d}
\end{figure}

Figure \ref{fig:dist} and \ref{fig:violin} illustrate the distributions and violin plots of chemical properties related to drug-likeness, solubility, and synthesizability. For the drug-likeness, most of the generated molecules distributed around 0.85, and the mean drug-likeness score is 0.78. Compared to the mean drug-likeness score (0.73) of training dataset, our model can generate relatively high score molecules, which can be credited trained on large dataset preserving more sub-structures. For the solubulity score, the distributions of generated and training molecules are similar, which indicated that our model can learn the solubility feature well. For the synthesizability score, most distributed in 0.85, and the mean scores of generated and training molecules are identical (0.69). It exhibits that even though our model can generate molecules with more sub-structures, the ease with which they can be synthesized is not compromised. These violin plots of generated and training molecules show similar shape except SA scores, which indicates that our model can sample molecules from the training distributions. 

\begin{figure}[t]
\centering
\includegraphics[width=0.7\textwidth]{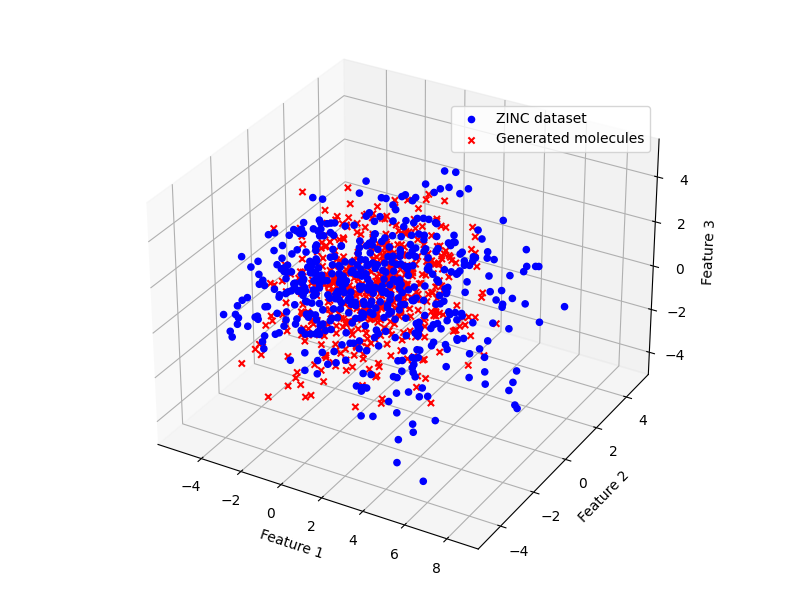}
\caption{Visualization of clustering generated and trained molecules in three-dimensional (3D) space.}
\label{fig:3d}
\end{figure}

We also perform further visualization analysis by clustering the training and generated molecules. Figures \ref{fig:2d} and \ref{fig:3d} show these visualization results in 2D and 3D, respectively. We feed the generated and trained molecules into the trained discriminator and extract the embeddings using principal component analysis \cite{mackiewicz1993principal} for dimension reduction. Finally, the dimensionally reduced embeddings are clustered for visualization. In general, the red scatters are distributed among the blue ones, which indicates that our model can sample from the train data distribution in terms of embeddings. It proves the learning ability of our model.

\section{Conclusion}
\label{sec:con}
This study presents a novel molecular GAN that leverages BPE tokenizer to tokenize SMILES strings. It integrates an RL algorithm to guide and stabilize training for \textit{de novo} molecular generation. The RL algorithm assists our GAN in effectively balancing exploration and exploitation during training, ensuring that the generated molecules adhere not only to the structural constraints encoded in SMILES representations but also exhibit desirable property distributions such as drug-likeness and synthesizability. Additionally, our GAN employs autoregressive predictions at the atomic level and incorporates innovative reward strategies. This approach enhances computational efficiency, thereby advancing the capabilities of \textit{de novo} molecular generation.

In future work, we plan to advance our GAN's capability to generate rare molecules by exploring more sophisticated reward mechanisms and training strategies. Additionally, scaling our approach to accommodate larger molecular libraries will be a focus, aiming to broaden the applicability and impact of our methodology in computational chemistry and drug discovery research.

\section*{Acknowledgements}
This research was partially supported by the International Research Fellow of Japan Society for the Promotion of Science (Postdoctoral Fellowships for Research in Japan [Standard]) and KAKENHI (20K11830), Japan.


\bibliographystyle{unsrt}
\bibliography{refs}

\begin{thebibliography}{10}

\bibitem{meyers2021novo}
Joshua Meyers, Benedek Fabian, and Nathan Brown.
\newblock De novo molecular design and generative models.
\newblock {\em Drug Discovery Today}, 26(11):2707--2715, 2021.

\bibitem{martinelli2022generative}
Dominic~D Martinelli.
\newblock Generative machine learning for de novo drug discovery: A systematic review.
\newblock {\em Computers in Biology and Medicine}, 145:105403, 2022.

\bibitem{medina2014balancing}
Jos{\'e}~L Medina-Franco, Karina Martinez-Mayorga, and Nathalie Meurice.
\newblock Balancing novelty with confined chemical space in modern drug discovery.
\newblock {\em Expert Opinion on Drug Discovery}, 9(2):151--165, 2014.

\bibitem{zhang2024cd}
Jinli Zhang, Shaomeng Wang, Zongli Jiang, Zhijie Chen, and Xiaolu Bai.
\newblock Cd-net: Cascaded 3d dilated convolutional neural network for pneumonia lesion segmentation.
\newblock {\em Computers in Biology and Medicine}, page 108311, 2024.

\bibitem{zhang2024quantitative}
Jinli Zhang, Zhenbo Wang, Zongli Jiang, Man Wu, Chen Li, and Yoshihiro Yamanishi.
\newblock Quantitative evaluation of molecular generation performance of graph-based gans.
\newblock {\em Software Quality Journal}, pages 1--29, 2024.

\bibitem{zhang2020hierarchy}
Jinli Zhang, Zongli Jiang, Yongping Du, Tong Li, Yida Wang, and Xiaohua Hu.
\newblock Hierarchy construction and classification of heterogeneous information networks based on rsdaef.
\newblock {\em Data \& Knowledge Engineering}, 127:101790, 2020.

\bibitem{zhang2019predicting}
Jinli Zhang, Xiaohua Hu, Zongli Jiang, Bo~Song, Wei Quan, and Zheng Chen.
\newblock Predicting disease-related rna associations based on graph convolutional attention network.
\newblock In {\em 2019 IEEE International Conference on Bioinformatics and Biomedicine (BIBM)}, pages 177--182. IEEE, 2019.

\bibitem{kusner2017grammar}
Matt~J Kusner, Brooks Paige, and Jos{\'e}~Miguel Hern{\'a}ndez-Lobato.
\newblock Grammar variational autoencoder.
\newblock In {\em International Conference on Machine Learning}, pages 1945--1954, 2017.

\bibitem{ma2021gf}
Changsheng Ma and Xiangliang Zhang.
\newblock {GF-VAE}: A flow-based variational autoencoder for molecule generation.
\newblock In {\em Proceedings of the 30th ACM International Conference on Information \& Knowledge Management}, pages 1181--1190, 2021.

\bibitem{vignacdigress}
Clement Vignac, Igor Krawczuk, Antoine Siraudin, Bohan Wang, Volkan Cevher, and Pascal Frossard.
\newblock {DiGress}: Discrete denoising diffusion for graph generation.
\newblock In {\em The Eleventh International Conference on Learning Representations}, 2022.

\bibitem{xu2023geometric}
Minkai Xu, Alexander~S Powers, Ron~O Dror, Stefano Ermon, and Jure Leskovec.
\newblock Geometric latent diffusion models for {3D} molecule generation.
\newblock In {\em International Conference on Machine Learning}, pages 38592--38610, 2023.

\bibitem{li2022transformer}
Chen Li, Chikashige Yamanaka, Kazuma Kaitoh, and Yoshihiro Yamanishi.
\newblock Transformer-based objective-reinforced generative adversarial network to generate desired molecules.
\newblock In {\em IJCAI}, pages 3884--3890, 2022.

\bibitem{li2023spotgan}
Chen Li and Yoshihiro Yamanishi.
\newblock {SpotGAN}: A reverse-transformer gan generates scaffold-constrained molecules with property optimization.
\newblock In {\em Joint European Conference on Machine Learning and Knowledge Discovery in Databases}, pages 323--338, 2023.

\bibitem{weininger1988smiles}
David Weininger.
\newblock {SMILES}, a chemical language and information system. 1. introduction to methodology and encoding rules.
\newblock {\em Journal of Chemical Information and Computer Sciences}, 28(1):31--36, 1988.

\bibitem{kearnes2016molecular}
Steven Kearnes, Kevin McCloskey, Marc Berndl, Vijay Pande, and Patrick Riley.
\newblock Molecular graph convolutions: Moving beyond fingerprints.
\newblock {\em Journal of Computer-Aided Molecular Design}, 30:595--608, 2016.

\bibitem{chowdhary2020natural}
KR1442 Chowdhary and KR~Chowdhary.
\newblock Natural language processing.
\newblock {\em Fundamentals of Artificial Intelligence}, pages 603--649, 2020.

\bibitem{sennrich2015neural}
Rico Sennrich, Barry Haddow, and Alexandra Birch.
\newblock Neural machine translation of rare words with subword units.
\newblock {\em arXiv preprint arXiv:1508.07909}, 2015.

\bibitem{liu2019roberta}
Yinhan Liu, Myle Ott, Naman Goyal, Jingfei Du, Mandar Joshi, Danqi Chen, Omer Levy, Mike Lewis, Luke Zettlemoyer, and Veselin Stoyanov.
\newblock {RoBERTa}: A robustly optimized bert pretraining approach.
\newblock {\em arXiv preprint arXiv:1907.11692}, 2019.

\bibitem{radford2019language}
Alec Radford, Jeffrey Wu, Rewon Child, David Luan, Dario Amodei, Ilya Sutskever, et~al.
\newblock Language models are unsupervised multitask learners.
\newblock {\em OpenAI blog}, 1(8):9, 2019.

\bibitem{guimaraes2017objective}
Gabriel~Lima Guimaraes, Benjamin Sanchez-Lengeling, Carlos Outeiral, Pedro Luis~Cunha Farias, and Al{\'a}n Aspuru-Guzik.
\newblock Objective-reinforced generative adversarial networks ({ORGAN}) for sequence generation models.
\newblock {\em arXiv preprint arXiv:1705.10843}, 2017.

\bibitem{li2024tengan}
Chen Li and Yoshihiro Yamanishi.
\newblock {TenGAN}: Pure transformer encoders make an efficient discrete gan for de novo molecular generation.
\newblock In {\em International Conference on Artificial Intelligence and Statistics}, pages 361--369, 2024.

\bibitem{williams1992simple}
Ronald~J Williams.
\newblock Simple statistical gradient-following algorithms for connectionist reinforcement learning.
\newblock {\em Machine Learning}, 8:229--256, 1992.

\bibitem{graves2012long}
Alex Graves and Alex Graves.
\newblock Long short-term memory.
\newblock {\em Supervised Sequence Labelling with Recurrent Neural Networks}, pages 37--45, 2012.

\bibitem{li2024gxvaes}
Chen Li and Yoshihiro Yamanishi.
\newblock {GxVAEs}: Two joint {VAE}s generate hit molecules from gene expression profiles.
\newblock In {\em Proceedings of the AAAI Conference on Artificial Intelligence}, volume~38, pages 13455--13463, 2024.

\bibitem{dai2018syntax}
Hanjun Dai, Yingtao Tian, Bo~Dai, Steven Skiena, and Le~Song.
\newblock Syntax-directed variational autoencoder for structured data.
\newblock In {\em International Conference on Learning Representations}, 2018.

\bibitem{tang2023macgan}
Huidong Tang, Chen Li, Shuai Jiang, Huachong Yu, Sayaka Kamei, Yoshihiro Yamanishi, and Yasuhiko Morimoto.
\newblock {MacGAN}: A moment-actor-critic reinforcement learning-based generative adversarial network for molecular generation.
\newblock In {\em APWeb-WAIM Joint International Conference on Web and Big Data}, pages 127--141, 2023.

\bibitem{browne2012survey}
Cameron~B Browne, Edward Powley, Daniel Whitehouse, Simon~M Lucas, Peter~I Cowling, Philipp Rohlfshagen, Stephen Tavener, Diego Perez, Spyridon Samothrakis, and Simon Colton.
\newblock A survey of monte carlo tree search methods.
\newblock {\em IEEE Transactions on Computational Intelligence and AI in Games}, 4(1):1--43, 2012.

\bibitem{tang2023earlgan}
Huidong Tang, Chen Li, Shuai Jiang, Huachong Yu, Sayaka Kamei, Yoshihiro Yamanishi, and Yasuhiko Morimoto.
\newblock {EarlGAN}: An enhanced actor--critic reinforcement learning agent-driven gan for de novo drug design.
\newblock {\em Pattern Recognition Letters}, 175:45--51, 2023.

\bibitem{bahdanau2022actor}
Dzmitry Bahdanau, Philemon Brakel, Kelvin Xu, Anirudh Goyal, Ryan Lowe, Joelle Pineau, Aaron Courville, and Yoshua Bengio.
\newblock An actor-critic algorithm for sequence prediction.
\newblock In {\em International Conference on Learning Representations}, 2022.

\bibitem{krenn2020self}
Mario Krenn, Florian H{\"a}se, AkshatKumar Nigam, Pascal Friederich, and Alan Aspuru-Guzik.
\newblock Self-referencing embedded strings ({SELFIES}): A 100\% robust molecular string representation.
\newblock {\em Machine Learning: Science and Technology}, 1(4):045024, 2020.

\bibitem{krenn2022selfies}
Mario Krenn, Qianxiang Ai, Senja Barthel, Nessa Carson, Angelo Frei, Nathan~C Frey, Pascal Friederich, Th{\'e}ophile Gaudin, Alberto~Alexander Gayle, Kevin~Maik Jablonka, et~al.
\newblock {SELFIES} and the future of molecular string representations.
\newblock {\em Patterns}, 3(10), 2022.

\bibitem{de2018molgan}
Nicola De~Cao and Thomas Kipf.
\newblock {MolGAN}: An implicit generative model for small molecular graphs.
\newblock {\em arXiv preprint arXiv:1805.11973}, 2018.

\bibitem{polsterl2021adversarial}
Sebastian P{\"o}lsterl and Christian Wachinger.
\newblock Adversarial learned molecular graph inference and generation.
\newblock {\em Machine Learning and Knowledge Discovery in Databases: European Conference, ECML PKDD 2020, Ghent, Belgium, September 14--18, 2020, Proceedings, Part II}, pages 173--189, 2021.

\bibitem{shigraphaf}
Chence Shi, Minkai Xu, Zhaocheng Zhu, Weinan Zhang, Ming Zhang, and Jian Tang.
\newblock {GraphAF}: A flow-based autoregressive model for molecular graph generation.
\newblock In {\em International Conference on Learning Representations}, 2020.

\bibitem{dinh2016density}
Laurent Dinh, Jascha Sohl-Dickstein, and Samy Bengio.
\newblock Density estimation using real {NVP}.
\newblock {\em arXiv preprint arXiv:1605.08803}, 2016.

\bibitem{you2018graph}
Jiaxuan You, Bowen Liu, Zhitao Ying, Vijay Pande, and Jure Leskovec.
\newblock Graph convolutional policy network for goal-directed molecular graph generation.
\newblock {\em Advances in Neural Information Processing Systems}, 31, 2018.

\bibitem{popova2019molecularrnn}
Mariya Popova, Mykhailo Shvets, Junier Oliva, and Olexandr Isayev.
\newblock {MolecularRNN}: Generating realistic molecular graphs with optimized properties.
\newblock {\em arXiv preprint arXiv:1905.13372}, 2019.

\bibitem{devlin-etal-2019-bert}
Jacob Devlin, Ming-Wei Chang, Kenton Lee, and Kristina Toutanova.
\newblock {BERT}: Pre-training of deep bidirectional transformers for language understanding.
\newblock In {\em Proceedings of NAACL-HLT}, pages 4171--4186, 2019.

\bibitem{achiam2023gpt}
Josh Achiam, Steven Adler, Sandhini Agarwal, Lama Ahmad, Ilge Akkaya, Florencia~Leoni Aleman, Diogo Almeida, Janko Altenschmidt, Sam Altman, Shyamal Anadkat, et~al.
\newblock Gpt-4 technical report.
\newblock {\em arXiv preprint arXiv:2303.08774}, 2023.

\bibitem{li2021smiles}
Xinhao Li and Denis Fourches.
\newblock Smiles pair encoding: a data-driven substructure tokenization algorithm for deep learning.
\newblock {\em Journal of chemical information and modeling}, 61(4):1560--1569, 2021.

\bibitem{ucak2023improving}
Umit~V Ucak, Islambek Ashyrmamatov, and Juyong Lee.
\newblock Improving the quality of chemical language model outcomes with atom-in-smiles tokenization.
\newblock {\em Journal of cheminformatics}, 15(1):55, 2023.

\bibitem{Kida1999BytePE}
Takuya Kida, Shuichi Fukamachi, Masayuki Takeda, Ayumi Shinohara, Takeshi Shinohara, and Setsuo Arikawa.
\newblock Byte pair encoding: a text compression scheme that accelerates pattern matching.
\newblock Technical Report DOI-TR-161, Department of Informatics, Kyushu University, 1999.

\bibitem{irwin2012zinc}
John~J Irwin, Teague Sterling, Michael~M Mysinger, Erin~S Bolstad, and Ryan~G Coleman.
\newblock Zinc: a free tool to discover chemistry for biology.
\newblock {\em Journal of chemical information and modeling}, 52(7):1757--1768, 2012.

\bibitem{srivastava2014dropout}
Nitish Srivastava, Geoffrey Hinton, Alex Krizhevsky, Ilya Sutskever, and Ruslan Salakhutdinov.
\newblock Dropout: a simple way to prevent neural networks from overfitting.
\newblock {\em The journal of machine learning research}, 15(1):1929--1958, 2014.

\bibitem{jin2018junction}
Wengong Jin, Regina Barzilay, and Tommi Jaakkola.
\newblock Junction tree variational autoencoder for molecular graph generation.
\newblock In {\em International Conference on Machine Learning}, pages 2323--2332, 2018.

\bibitem{mackiewicz1993principal}
Andrzej Ma{\'c}kiewicz and Waldemar Ratajczak.
\newblock Principal components analysis (pca).
\newblock {\em Computers \& Geosciences}, 19(3):303--342, 1993.

\end{thebibliography}
\end{document}